\documentclass[11pt]{article}

\usepackage[final]{acl}

\usepackage{times}
\usepackage{latexsym}
\usepackage{amsmath}
\usepackage{booktabs}
\usepackage[T1]{fontenc}

\usepackage[utf8]{inputenc}
\usepackage{tikz}
\usetikzlibrary{positioning,arrows.meta,calc,fit,backgrounds,shapes.geometric}
\usepackage{microtype}

\usepackage{inconsolata}

\usepackage{graphicx}

\title{Multimodal Item Parameter Estimation using Simulated Response Probabilities}

\author{Christopher Ormerod \\
	\texttt{College Board}\\
	\texttt{cormerod@collegeboard.org} \\\And
	YoungKoung Kim \\
	\texttt{College Board}\\
	\texttt{ykim@collegeboard.org} }

\begin{document}
	\maketitle
	\begin{abstract}
		We present results from reconstructing multiple-choice model (MCM) and three-parameter logistic (3PL) model curves using a fine-tuned multimodal large language model (LLM) based on Qwen3.5. The model is prompted and fine-tuned to replicate choice probabilities across a large training corpus of multiple-choice items containing both image and text stimuli, conditioned on a labeled set of student ability levels. By learning to reproduce the systematic error patterns of students across a discrete range of abilities, the LLM implicitly captures the underlying response probabilities encoded in the 3PL and MCM curves. This allows us to accurately approximate item difficulty on a held-out test set directly from the model's predicted option probabilities.
	\end{abstract}
	
	\section{Introduction}
	
	Item difficulty parameters are essential for evaluating whether an assessment accurately measures student knowledge, differentiating between students, and identifying specific learning gaps. The field-testing required to obtain these parameters is not only expensive and labor-intensive, but also poses an inherent security risk. For these reasons, researchers have explored machine learning techniques to approximate these parameters \cite{alkhuzaey_text-based_2024}. Given the capabilities of Large Language Models (LLMs), a growing body of literature has emerged on their use in predicting item difficulty parameters. Two characteristically distinct approaches have been proposed: treating the question stimulus and its features as input to a fine-tuned LLM in a regression framework \cite{li_item_2025}, and using LLMs as simulated respondents to test stimuli \cite{maeda_field-testing_2025}.
	
	Generative LLMs have become increasingly powerful due to advances in training regimes \cite{wang_reinforcement_2024} and the ability to scale the transformer architecture \cite{vaswani_attention_2017} to the point at which emergent capabilities arise \cite{wei_emergent_2022}. The dominant paradigm for generative models has shifted from open-ended text generation to instruction-following via prompting \cite{chung_scaling_2022}. The flexibility of prompting allows researchers to instruct LLMs to answer questions in targeted ways, enabling the simulation of students with specified combinations of skills \cite{lu_generative_2024}, knowledge \cite{he-yueya_psychometric_2024}, and ability levels \cite{liu_leveraging_2025}. The present study extends work in which models are fine-tuned in a parameter-efficient manner to simulate students of varying ability \cite{scarlatos_smart_2025}, subsequently applied to multiple-choice questions \cite{ormerod_reconstructing_2026}.
	
	Prior work has examined the reconstruction of item characteristic curves \cite{ormerod_reconstructing_2026} using the Nominal Response Model (NRM) and the Two-Parameter Logistic (2PL) model to estimate item parameters \cite{thissen_taxonomy_1986}. Both models share the limiting assumption that the probability of a correct response approaches zero as ability decreases. The Three-Parameter Logistic (3PL) model \cite{thissen_taxonomy_1986} and the Multiple-Choice Model (MCM) \cite{thissen_response_1984, samejima_estimation_1969} address this by incorporating non-zero lower asymptotes, thereby accounting for the possibility of guessing. First, the present work introduces several improvements to the framework of \cite{ormerod_reconstructing_2026} and applies it to a large corpus of mathematics items. Second, we demonstrate this framework for items with both text and images by fine-tuning multimodal models with hybrid transformer-based architectures. While this may appear rather cumbersome, we are able to reproduce the item difficulty parameter with remarkably high accuracy compared to baseline regression-based methods.
	
	We organize this paper as follows: Our method section, \S \ref{sec:method}, covers the nature of the response and item data used in this project, the response modeling applied to determine item parameters, the way in which we discretize ability levels, and the way we use fine-tuned generative LLMs, the baseline approaches, and the metrics used for evaluation. In \S \ref{sec:results}, we present our results, before concluding with a discussion in \S \ref{sec:discussion}.
	
	\section{Method}\label{sec:method}
	
\begin{figure}[!ht]
		\begin{tikzpicture}[xscale=0.7]
			\node[draw=black, rounded corners=3pt, fill=green!5](r) at (0,0) {Responses};
			\node[draw=black, rounded corners=3pt, fill=blue!5](pl) at (2.5,1) {3PL};
			\node[draw=black, rounded corners=3pt, fill=blue!5](mcm) at (2.5,-1) {MCM};
			\node[draw=black, rounded corners=3pt, fill=blue!5](dmcm) at (5,0) {\begin{tabular}{c}Discrete\\MCM \end{tabular}};
			\node[draw=black, rounded corners=3pt, fill=red!5](llm) at (8,-1) {LLM};
			\draw[very thick, <->, rounded corners=10pt] (dmcm) -| (llm);
			\draw[very thick, ->, rounded corners=10pt] (r) |- (pl);
			\draw[very thick, ->, rounded corners=10pt] (r) |- (mcm);
			\draw[very thick, <->, rounded corners=10pt] (mcm) -| (dmcm);
			\draw[very thick, ->, rounded corners=10pt] (dmcm) |- (pl);
			\node at (-.3,1) {$i$};
			\node at (-.3,-1) {$ii$};
			\node at (5.7,-1) {$iii$, $iv$};
			\node at (5.5,1) {$v$};
			\node at (8,0.3) {$vi$,$vii$};
		\end{tikzpicture}
		\caption{Diagram of the relationships between the data and models used in this study.\label{fig:models}}
	\end{figure}
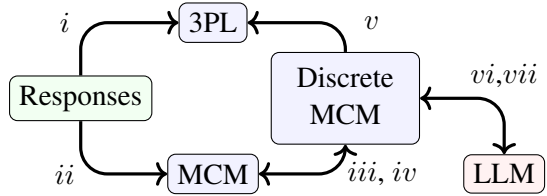

	To explain our method, we have raw data in the form of responses, and various models based on that data: 3PL models \cite{thissen_taxonomy_1986}, MCMs \cite{thissen_response_1984}, the discrete MCMs \cite{ormerod_reconstructing_2026}, and LLMs \cite{touvron_llama_2023}. These models have been represented in Figure \ref{fig:models}, while the arrows, labeled $i$ to $vii$, denote the various ways to define parameters of each model either in terms of the raw responses or other models.
	
	The following subsections are devoted to defining the data used in this study, the models, and the method of deriving the parameters of each model. With all these defined, our goal is to reconstruct the parameters of the 3PL models and MCMs defined by $i$ and $ii$, respectively.
	
	\subsection{Data}
	
	The raw data in this study consists of two distinct objects: items and responses to those items.
	
	\subsubsection{Items}
	
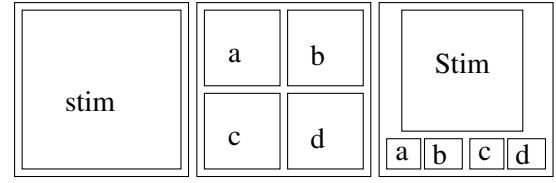
\begin{figure}[!ht]
		\centering
		\begin{tikzpicture}
			\draw(0,0) rectangle (2.3,2.3);
			\draw(.1,.1) rectangle (2.2,2.2);
			\node at (1,1) {stim};
		\end{tikzpicture}
		\begin{tikzpicture}
			\draw(0,0) rectangle (2.3,2.3);
			\draw (.1,.1) rectangle (1.1,1.1);
			\draw (1.2,.1) rectangle (2.2,1.1);
			\draw (1.2,1.2) rectangle (2.2,2.2);
			\draw (0.1,1.2) rectangle (1.1,2.2);
			\node at (.5,1.6) {a};
			\node at (1.6,1.6) {b};
			\node at (.5,.5) {c};
			\node at (1.6,0.5) {d};
		\end{tikzpicture}
		\begin{tikzpicture}
			\draw(0,0) rectangle (2.3,2.3);
			\draw(.1,.1) rectangle (.55,0.5);
			\draw(.6,.1) rectangle (1.1,0.5);
			\draw(1.2,.1) rectangle (1.65,0.5);
			\draw(1.7,.1) rectangle (2.2,0.5);
			\draw(.3,.6) rectangle (1.9,2.2);
			\node at (0.3,0.3) {a};
			\node at (0.8,0.3) {b};
			\node at (1.4,0.3) {c};
			\node at (1.9,0.3) {d};
			\node at (1.1, 1.5) {Stim};
		\end{tikzpicture}
		\caption{The way in which stimulus and option images were combined into a single image. The left image depicts stimulus-only images, the middle depicts option-only images, and the right depicts when both stimulus and options are present.}
		\label{fig:im-combine}
	\end{figure}

	Each item consists of a stimulus and a set of four distinct options. Since our understanding of Mathematics can be inherently visual, we allow both the stimulus and the options to have an associated image component. In order to facilitate this combination of images as input, we transform the images into one image in the manner presented in Figure \ref{fig:im-combine}.
	
	This dataset is produced using item models, which specify the structure, content, and variable features needed to produce new, functional test items \cite{gierl_automatic_2013}. There were a total of 970 item models and a total of 4,848 items. In order to avoid conflation, we split the items into training, development, and test sets by item model. This gave a natural split at the item level represented in Table \ref{tab:split}.
	
	\subsection{Response Modeling}
	
	There are 13.88 million responses. The ability level of each respondent was independently derived from a larger calibrated dataset of items. These responses provide us with an average of approximately 2860 responses per item with a minimum of 860 responses. The average response ability level was 0.31 and the standard deviation was approximately 1. 
	\begin{table}
		\begin{tabular}{l r r r}\toprule
			&Item && \\
			& Model &  Items & Responses\\ \midrule
			Train  & 727 & 3633 &  $10.2 \times 10^6$\\
			Development & 97 & 485 & $1.42\times 10^6$\\
			Test & 146 & 730 & $2.24\times 10^6$\\ \midrule
			Total & 970 & 4848 & $13.88\times 10^6$\\ \bottomrule
		\end{tabular}
		\caption{A representation of how the items were split into test, train, and development. \label{tab:split}}
	\end{table}
	
	Given that each of these responses can be associated with a particular value of student ability, we can model the probability that a student $j$, with ability level $\theta_j$, provides a correct answer to item $i$ by
	\begin{equation}\label{eq:3pl}
		P_{ij}(X = 1) = c_i + (1-c_i) \sigma( a_{i}(\theta_j - b_i)) 
	\end{equation}
	where $\sigma$ is the usual sigmoid function. There are at least 10 correct or incorrect responses to any particular item, facilitating stable estimates of the item parameters, $\{ a_i, b_i, c_i\}$ by minimizing the negative log likelihood function using the Limited-memory Broyden–Fletcher–Goldfarb–Shanno with Bound constraints (L-BFGS-B).
	
	To satisfy the conditions that $a_i > 0$ and $0 < c_i < 1$, we let $a_i = e^{\alpha_i}$ and $c_i = \sigma(\gamma_i)$. Then our first arrow, $i$, is provided by the optimization 
	\begin{align}
		&a_i, b_i, c_i = \mathrm{argmax} \left\{\sum_j [y_j \log P_{ij}\right.\nonumber\\
		& \left. + (1-y_j) \log (1- P_{ij}) ] - \lambda (\alpha_i^2 + b_i^2)\right.\Bigg\} \tag{$i$}
	\end{align}
	where the first two terms are the usual negative log likelihood function and the third term is a small $L2$ regularization term with $\lambda = 10^{-4}$. We used the Limited-memory Broyden–Fletcher–Goldfarb–Shanno (LBFGS) method.
	
	The model described by \eqref{eq:3pl} is known as the three-parameter logistic model (3PL) \cite{thissen_taxonomy_1986}. We call $a_i$ the discrimination parameter, $b_i$ the difficulty parameter, and $c_i$ is called the guessing parameter. It has the asymptotic property that as $\theta_j$ is large and negative, the probability of obtaining the correct response is given by random chance; hence, it is typically assumed to be close to the reciprocal of the number of options. As $c_i \to 0$, we obtain the two-parameter logistic model \cite{thissen_taxonomy_1986}.
	
	The 3PL model is useful in defining a difficulty parameter; however, it does not allow us to provide any details regarding the strength of distractor items. Given options $k = 1, \ldots, K$, the probability of a person of ability level $\theta_j$ providing answer $k$ to item $i$ is modeled by
	\begin{equation}
		P_{ijk} = P_{ij}(X = k) = \dfrac{\exp(\eta_{ijk}) + d_{ik}}{1 + \sum_k \exp(\eta_{ijk})},\label{eq:mcm}
	\end{equation}
	where $\eta_{ijk} = a_{ik} \theta_j + b_{ik}$. Given that $\sum_k P_{ij}(X =k) = 1$ and $d_{ik} > 0$, this provides us with $\sum_{k} d_{ik} = 1$ for any item. This model is a reformulation of the multiple-choice model from Thissen and Steinberg \shortcite{thissen_response_1984}. Alternatively, this generalizes the version of Samejima in which $d_{ik} = K^{-1}$ \shortcite{samejima_estimation_1969}. In a similar manner to the 3PL model, we may derive the values of $\{a_{ik}, b_{ik}, d_{ik}\}$ using the optimization
	\begin{align}
		&\{a_{ik}, b_{ik}, d_{ik}\} = \mathrm{argmax} \left\{ \sum_{j=1} \log P_{ijk}   \right. \nonumber \\
		& \left. - \lambda (a_{ik}^2 + b_{ik}^2) \right\}  \tag{$ii$}
	\end{align}
	with $\lambda = 10^{-4}$ using LBFGS. This provides us with the arrow $ii$. In combination with $i$, this provides the target values for our modeling.
	
	\subsection{Discrete Ability Modeling}
	
	The main idea from \cite{ormerod_reconstructing_2026} is that we are binning the values of $\theta_j$ into $L$ categories. The cutoff points for each category are denoted $\gamma_i$ so that $I_l = (\gamma_{l-1}, \gamma_l]$ with 
	\[
	-\infty = \gamma_0 < \gamma_1 < \ldots < \gamma_L = \infty.
	\]
	This provides us with intervals $I_1, \ldots, I_L$ such that each $\theta_j \in I_{l_j}$. The greater the $L$, the more fine-grained the division of abilities. Each interval is associated with a loosely descriptive label. The labels and intervals are presented in Table \ref{tab:labels}.
	
	\begin{table}[!ht]
		\begin{center}
			\begin{tabular}{l rr} \toprule
				Label & $\gamma_{i-1}$ & $\gamma_i$ \\ \midrule
				Foundational            & $-\infty$ & -2.7\\
				Beginning               & -2.7 & -2.4 \\
				Emerging foundations    & -2.4 & -2.1\\
				Early developing        & -2.1 & -1.8\\
				Developing              & -1.8 & -1.5\\
				Developing proficiency  & -1.5 & -1.2\\
				Approaching basic       & -1.2 & -0.9\\
				Basic                   & -0.9 & -0.6\\
				Basic plus              & -0.6 & -0.3\\
				Approaching average     & -0.3 & 0\\
				Average                 & 0     & 0.3\\
				Average plus            & 0.3   & 0.6\\
				Above average           & 0.6   & 0.9\\
				Strong                  & 0.9   & 1.2\\
				Very strong             & 1.2   & 1.5\\
				Advanced                & 1.5   & 1.8\\
				Highly advanced         & 1.8   & 2.1\\
				Exceptional             & 2.1   & 2.4\\
				Outstanding             & 2.4   & 2.7\\
				Elite mastery           & 2.7 & $\infty$\\ \bottomrule
			\end{tabular}
		\end{center}
		\caption{A list of the descriptive labels used to calibrate the language model. \label{tab:labels}}
	\end{table}
	
	Following the work of \cite{ormerod_reconstructing_2026}, for item $i$, we associate each interval with a probability value for each option, which gives us a function
	\begin{equation}
		f_i(I_j) = (\rho_{ij1}, \ldots, \rho_{ijK}).
	\end{equation}
	This means we obtain, as data, an $L\times K$ matrix of probability values.  
	
	Our underlying prior is that the population is normally distributed such that $\{\theta_j\} \sim \mathcal{N}(\mu, \varsigma^2)$. We find it convenient to state functions associated with this distribution explicitly. The probability and cumulative distribution functions (PDF \& CDF), $f(\theta)$ and $F(\theta)$ are given by 
	\begin{align*}
		f(\theta) = \frac{1}{\varsigma}\phi\left(\frac{\theta-\mu}{\varsigma}\right), \,\, F(\theta) = \Phi\left(\frac{\theta-\mu}{\varsigma}\right),
	\end{align*}
	where
	\begin{align*}
		&\phi(z) = \dfrac{1}{\sqrt{2\pi}} e^{-\frac{z^2}{2}}, \,\, \Phi(z) = \int_{-\infty}^{z} \phi(t) \mathrm{d}t,
	\end{align*}
	It makes sense to define the conditional probability that $\theta \in I_j$ as
	\[
	w_j = F(\gamma_j) - F(\gamma_{j-1})
	\]

	In \cite{ormerod_reconstructing_2026}, each interval was associated with the expected value under this prior, with $\rho_{ijk}$ defined as the evaluation of \eqref{eq:mcm} at this value. It actually makes more sense to associate the interval with the expected value of the function itself. Provided that $P_{ijk}$ is defined by \eqref{eq:mcm}, then the definition of $\rho_{ijk}$ may be determined by
	\begin{align}
		&\rho_{ijk} = \frac{1}{w_j} \int_{\gamma_{j-1}}^{\gamma_j} f(\theta) P_{i\theta k} \mathrm{d}\theta, \tag{$iii$}
	\end{align}
	where $P_{i\theta k}$ is the function of $\theta$ rather than the particular value at $\theta_j$. This provides us with $iii$. 
	
	Furthermore, given that these are continuous values rather than discrete values, we can approximate the reconstruction of the MCM by optimizing 
	\begin{align*}
		(a_{ik},& b_{ik},d_{ik}) = \mathrm{argmin}\Bigg( \sum_j w_j \Bigg( \rho_{ijk} -\\
		& \frac{1}{w_{j}}\int_{\gamma_{j-1}}^{\gamma_j} f(\theta) P_{i\theta k} \mathrm{d}\theta\Bigg)^2 \Bigg). \tag{$iv$}
	\end{align*}
	Alternatively, we can specialize the value of $k$ to be the correct option, $\tilde{k}$, in which case we obtain the probability of being correct, which provides a viable path to reconstructing the parameters of \eqref{eq:3pl}. For completeness, we write this as
	\begin{align}
		(a_i, b_i&, c_i) = \mathrm{argmin} \Bigg( \sum_j w_j \Bigg( \rho_{ij\tilde{k}}  \nonumber \\
		& - \frac{1}{w_j}\int_{\gamma_{j-1}}^{\gamma_j} f(\theta) P_{i\theta} \mathrm{d}\theta \Bigg)^2 \Bigg) \tag{$v$}
	\end{align}
	where $P_{i\theta}$ is the model \eqref{eq:3pl} where the dependency on $\theta_j$ is replaced by the general function of $\theta$. This provides us with all the arrows that do not depend on the language model. 
	
	It is worth noting that these approximations, at each arrow, are not exact. For example, the arrow $i$ provides a mapping from the set of empirically observed responses to the IRT parameters; however, there is an underlying assumption that \eqref{eq:3pl} is a good fit for the empirical data. We assume that there are some regressive tendencies in which the parameters are linearly related to the final parameters. We use the development set to appropriately and linearly model this relationship, which does not change the overall Pearson correlations, but does affect the mean squared error. 
	
	\subsection{Language Modeling}
	
	The Qwen model series continues to provide the research community with a suite of excellent fine-tunable models for research purposes. The latest series of models, Qwen3.5, provides us with the opportunity to demonstrate the abilities of multimodal fine-tuning. The models we conservatively chose are the 4-billion- and 9-billion-parameter variants of Qwen3.5.
	
	\subsubsection{Parameter-efficient fine-tuning}
	
	These models are small enough to fit on a local machine with sufficient resources. In order to train these models, we employ parameter-efficient methods \cite{xu_parameter-efficient_2023} such as Low-Rank Adaptation (LoRA) \cite{hu_lora_2021} or LoRA with Quantization \cite{dettmers_qlora_2023} to effectively tune the model, requiring some knowledge of the underlying model structure. 
	
	The Qwen3.5 series not only integrates multimodal inputs but is also a hybrid structure similar to the Jamba series \cite{lieber_jamba_2024}. Each layer replaces multiheaded attention in the transformer architecture \cite{vaswani_attention_2017} with two distinct types of layers: Gated DeltaNet layers \cite{yang_gated_2025} and Gated Attention layers \cite{qiu_gated_2025}. The Gated DeltaNet layers have a structure similar to that of the Mamba2 layers in that the information is stored in hidden states and preserved through a selective gating mechanism. This means that the Gated DeltaNet layers have linear complexity while the Gated Attention layers have quadratic complexity, making this closer to a Jamba model than previous Qwen models. 
	
	\begin{figure}[!ht]
		\begin{tikzpicture}[yscale=1]
			\node[draw=black, rounded corners=2pt, fill=blue!10] at (0.1,-0.1) {Linear};
			\node[draw=black, rounded corners=2pt, fill=blue!10](l1) at (0,0) {Linear};
			\node[draw=black, rounded corners=2pt, fill=blue!10](l2) at (2,0) {Linear};
			\node[draw=black, rounded corners=2pt, fill=blue!10] at (4.1,-.1) {Proj.};
			\node[draw=black, rounded corners=2pt, fill=blue!10](l3) at (4,0) {Proj.};
			\node[draw=black, rounded corners=2pt, fill=blue!10](l4) at (6,0) {Linear};
			\node[draw=black, rounded corners=2pt, fill=red!10] at (0.1,0.7) {Conv.};
			\node[draw=black, rounded corners=2pt, fill=red!10](c1) at (0,.8) {Conv.};
			\node[draw=black, rounded corners=2pt, fill=red!10](c2) at (2,.8) {Conv.};
			\node[draw=black, rounded corners=2pt, fill=pink!10] at (0.1,2.0) {L2};
			\node[draw=black, rounded corners=2pt, fill=pink!10](L2) at (0,1.9) {L2};
			\node[draw=black, rounded corners=2pt, fill=green!10, minimum width=5.5cm](delta) at (2,3) {Gated Delta Rule};
			\node[draw=black, rounded corners=2pt, fill=purple!10, minimum width=3.5cm](norm) at (2,3.7) {Zero Centered RMSNorm};
			\node[draw=black, rounded corners=2pt, fill=blue!10](l5) at (2,5) {Linear};
			\draw[very thick] (l1) -- (c1);
			\draw[very thick] (l2) -- (c2);
			\draw[very thick] (c1) -- (L2);
			\draw[very thick] (L2.north) -- (L2.north |- delta.south);
			\draw[very thick] (c2.north) -- (c2.north |- delta.south);
			\draw[very thick] (l3.north) -- (l3.north |- delta.south);
			\draw[very thick] (delta) -- (norm);
			\draw[very thick] (norm) -- (l5);
			\draw[very thick,rounded corners=6pt] (l4) |- (2,4.3);
			\node[fill=white, draw=black,rounded corners=6pt] at (2,4.3) {${}_{\otimes}$};
			\node[fill=white, draw=black,rounded corners=6pt] at (0,1.35) {${}_{\sigma'}$};
			\node[fill=white, draw=black,rounded corners=6pt] at (2,1.35) {${}_{\sigma'}$};
			\node[fill=white, draw=black,rounded corners=6pt] at (6,1.35) {${}_{\sigma'}$};
			\draw[very thick] (2,-1) -- (l2);
			\draw[very thick,rounded corners=6pt] (2,-0.5) -| (l1);
			\draw[very thick,rounded corners=6pt] (2,-0.5) -| (l3);
			\draw[very thick,rounded corners=6pt] (2,-0.5) -| (l4);
			\node at (-.2,2.5) {$q$};
			\node at (.2,2.5) {$k$};
			\node at (-.2,2.5) {$q$};
			\node at (4.3,4.8) {Gated DeltaNet};
		\end{tikzpicture}
		\caption{The structure of the Gated DeltaNet layers. Here, $\sigma'$ is the SiLU function.\label{fig:gated}}
	\end{figure}
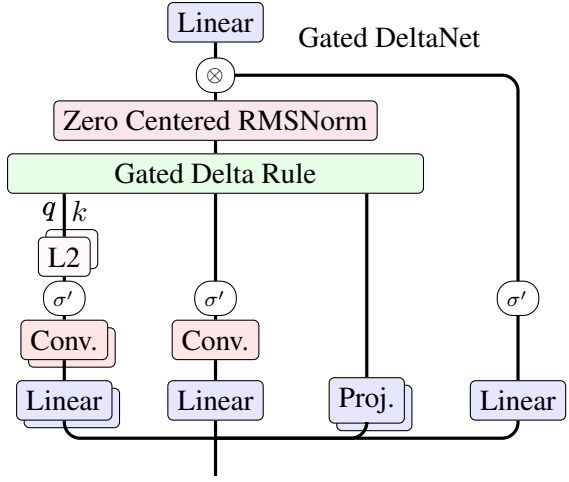
	
\begin{figure}[!ht]
		\begin{tikzpicture}[yscale=1]
			\node[draw=black, rounded corners=2pt, fill=blue!10](l1) at (0,0) {Linear};
			\node[draw=black, rounded corners=2pt, fill=blue!10](l2) at (2,0) {Linear};
			\node[draw=black, rounded corners=2pt, fill=blue!10](l3) at (4,0) {Linear};
			\node[draw=black, rounded corners=2pt, fill=blue!10](l4) at (6,0) {Linear};
			\node[draw=black, rounded corners=2pt, fill=red!10](n1) at (0,.8) {{\tiny \begin{tabular}{c}
						Zero-centered\\
						RMSNorm
			\end{tabular}}};
			\node[draw=black, rounded corners=2pt, fill=red!10](n2) at (2,.8) {{\tiny\begin{tabular}{c}
						Zero-centered\\
						RMSNorm
			\end{tabular}}};
			\node[draw=black, rounded corners=2pt, fill=pink!10](r1) at (0,1.7) {{\tiny \begin{tabular}{c}Partial \\ RoPE\end{tabular}}};
			\node[draw=black, rounded corners=2pt, fill=pink!10](r2) at (2,1.7) {{\tiny \begin{tabular}{c}Partial \\ RoPE\end{tabular}}};
			\node[draw=black, rounded corners=2pt, fill=green!10, minimum width=5.5cm](attn) at (2,3) {Scaled dot product attention};
			\node[draw=black, rounded corners=2pt, fill=blue!10](l5) at (2,4.3) {Linear};
			\draw[very thick] (l1) -- (n1);
			\draw[very thick] (l2) -- (n2);
			\draw[very thick] (n1) -- (r1);
			\draw[very thick] (n2) -- (r2);
			\draw[very thick] (l3) -- (l3.north |- attn.south);
			\draw[very thick] (r1) -- (r1.north |- attn.south);
			\draw[very thick] (r2) -- (r2.north |- attn.south);
			\draw[very thick] (2,-1) -- (l2);
			\draw[very thick,rounded corners=6pt] (2,-0.5) -| (l1);
			\draw[very thick,rounded corners=6pt] (2,-0.5) -| (l3);
			\draw[very thick,rounded corners=6pt] (2,-0.5) -| (l4);
			\draw[very thick,rounded corners=6pt] (2,3.7) -| (l4);
			\draw[very thick] (attn) -- (l5);
			\node[fill=white, draw=black,rounded corners=6pt] at (2,3.7) {${}_{\otimes}$};
			\node[fill=white, draw=black,rounded corners=6pt] at (6,1.35) {${}_{\sigma}$};
			\node at (-.2,2.4) {$q$};
			\node at (2-.2,2.4) {$k$};
			\node at (4-.2,2.4) {$v$};
			\node at (4.3,4.3) {Gated Attention};
		\end{tikzpicture}
		\caption{The structure of the Gated Attention layers. Here, $\sigma$ is the sigmoid function. \label{fig:structure}}
	\end{figure}
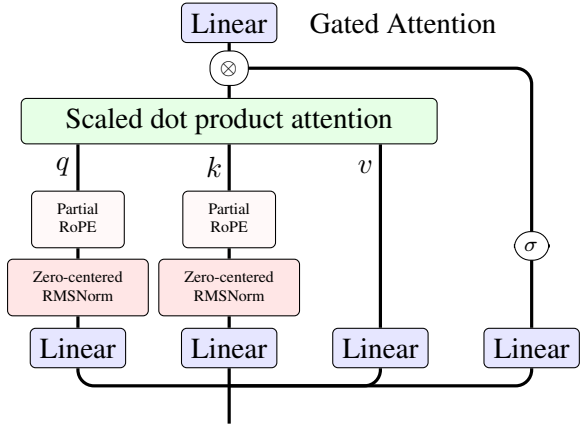

	Figure \ref{fig:structure} presents the internal structure of the Gated DeltaNet and Attention layers.

	The application of LoRA takes a selection of linear layers, of the form
	\[
	L(x) = Mx + b
	\]
	where $M$ is an $m\times n$ matrix, and replaces those linear layers with 
	\[
	\tilde{L}(x) = (M + BA)x + b
	\]
	where $A$ is an $r \times n$ matrix and $B$ is an $m \times r$ matrix with $r << \min(m,n)$. The main idea is that we freeze all other layers and keep $A$ and $B$ trainable \cite{hu_lora_2021}. This process has been very successful in automated scoring \cite{ormerod_automated_2024}, standards alignment \cite{han_fine-tuning_2026}, and item parameter estimation \cite{ormerod_reconstructing_2026}.
	
	Given instabilities in the training process encountered in the literature, we conservatively applied LoRA to the attention components, knowing this would essentially affect only a quarter of the hidden layers that would be typically used to fine-tune models. We would need a better understanding of how to fine-tune the Gated DeltaNet layers before considering the other layers. 
	
	\subsubsection{Prompting}
	
	The creation of a dataset that encodes information regarding student options requires formatting the data as a sequence of chats. Each chat consists of three components: a system text indicating the behavior of the model, a user text indicating a task to be completed, and an assistant text indicating how the task should be completed. These texts are functions of the ability level (\verb|ability_level|), the correct answer (\verb|correct|), the item text (\verb|stim_text|), and the option text (\verb|option_text|), which contains an enumeration of the options in text form. These are presented below:
	
	\vspace{.3cm}
	
	\rule{\linewidth}{0.4pt}
	\begin{verbatim}
		System: You are a student who must answer 
		either A, B, C, D in a multiple choice 
		question.
		
		**Ability Level**: {ability_level}.
		
		**Correct Answer**: {correct}
	\end{verbatim}
	\rule{\linewidth}{0.4pt}
	\begin{verbatim}
		User: Answer the following **Question** 
		as if you were a {ability_level} student.
		
		**Question**
		{stim_text}
		
		**Options**
		{option_text}
	\end{verbatim}
	\rule{\linewidth}{0.4pt}
	\begin{verbatim}
		Assistant: **Answer**: 
	\end{verbatim}
	\rule{\linewidth}{0.4pt}
	
	\vspace{.3cm}
	
	We leave the final text of the student option blank, truncating the text so that it does not satisfy the traditional requirements of a full chat. The idea is that we are obtaining a vector of next-token probabilities, which we will denote $v$ where the dimension of $v$ is the number of tokens in the model vocabulary. For convenience, we will assume the softmax has been applied so that each component of $v$, denoted $v_{\iota}$, is interpreted as a token probability.
	
	There are indices, $\iota_1, \ldots, \iota_K$, associated with each option. This allows us to construct the Mean Squared Error (MSE) loss function, defined by
	\begin{equation}
		\mathcal{L} = \sum_k \frac{1}{K}\left( v_{\iota_k} - \rho_{ijk} \right)^2,
	\end{equation}
	where $i$ denotes the item index and $j$ denotes the ability level index. Using the standard Adam optimizer with a weight decay mechanism and a learning rate of $10^{-5}$ over 10 epochs provides us with the training regime used to define the fine-tuned language model, essentially providing the arrow $vi$. Finally, inference using the prompt above provides the arrow $vii$.
	
	\subsubsection{Baseline models}
	
	Our baseline approach is to use the question stimulus and options as input and then treat the IRT parameters as targets for a regression problem. This means that we take an encoder-only transformer-based language model, such as MetaMath \cite{yu_metamath_2024} and MathBERT \cite{peng_mathbert_2021}, replace the head with a linear layer with one target, and use the MSE loss function, similar to above, with the outputs of the language model and the target IRT parameters as the arguments. This approach is the one that is most common in the literature \cite{bulut_item_2024, peters_text-based_2025}.
	
	Table~\ref{tab:models} summarizes the models used in this study, together with
	their approximate parameter counts and references.
	
	\begin{table*}[!ht]
		\centering
		\begin{tabular}{l l l l} \toprule
			\textbf{Model} & \textbf{Role} & \textbf{Params} & \textbf{Reference} \\ \midrule
			MathBERT     & Baseline    & 110M       & \cite{peng_mathbert_2021} \\
			MetaMath     & Baseline    & 70B         & \cite{yu_metamath_2024} \\
			Qwen3.5-4B   & Proposed & 4B         & \cite{qwen_team_qwen35_2026} \\
			Qwen3.5-9B   & Proposed & 9B         & \cite{qwen_team_qwen35_2026} \\ \bottomrule
		\end{tabular}
		\caption{Models used in this study, with approximate parameter counts and
			references. MathBERT \cite{peng_mathbert_2021} is built on BERT-base; MetaMath is a LLaMA-2
			fine-tune available at 7B, 13B, and 70B. \label{tab:models}}
	\end{table*}
	
	\subsection{Metrics}
	
	In reference to the literature on item parameter prediction, there are two dominant metrics used to compare the performance of difficulty prediction models: Pearson correlation and root mean squared error \cite{alkhuzaey_text-based_2024}.
	
	We would like to highlight another approach that fits nicely in the context of our study: a measure of the agreement under a discretization of difficulty. In communicating difficulty to educators, item difficulty parameters are often difficult to interpret. For this reason, what is often conveyed is a descriptive label that is similar to the descriptive labels we used to describe ability levels. The labels, as functions of the $b$ parameter, are presented in Table \ref{tab:diff_labs}. Under these descriptive labels, it is useful to obtain a measure of the agreement between the predicted labels and the calibrated labels. For this, we use the quadratic weighted kappa (QWK).

	\begin{table}[!ht]
		\centering
		\begin{tabular}{llll}
			\toprule
			\textbf{Label} & \textbf{Full Name} & \textbf{Condition} \\
			\midrule
			VH  & Very Hard        & $b \geq 0.842$ \\
			H   & Hard             & $0.253 \leq b < 0.842$ \\
			M   & Medium           & $-0.253 \leq b < 0.253$ \\
			E   & Easy             & $ -0.842 \leq b < -0.253$ \\
			VE  & Very Easy        & $b < -0.842$ \\
			\bottomrule
		\end{tabular}
		\caption{A list of the descriptive labels used to communicate difficulty. \label{tab:diff_labs}}
	\end{table}
	
	\section{Results} \label{sec:results}
	
	\subsection{Difficulty Prediction}
	
	Difficulty is the parameter of greatest practical interest, since it is the
	quantity most often reported to educators and used when assembling test forms.
	Table~\ref{tab:results} reports the Pearson correlation and root mean squared
	error (RMSE) between the predicted and calibrated parameters.
	
	\begin{table}[!ht]
		\centering
		\begin{tabular}{l | c c c} \toprule
			&  \multicolumn{3}{c}{Pearson} \\ 
			& a & b & c  \\\midrule
			MathBERT & 0.35 & 0.68 & 0.25 \\
			MetaMath & 0.34 & 0.75 & 0.28 \\
			
			Qwen3.5-4B & 0.26 & 0.80 & 0.45\\
			Qwen3.5-9B & 0.31 & 0.85 & 0.48  \\\midrule
			&  \multicolumn{3}{c}{RMSE} \\ \midrule
			
			MathBERT & 1.03 & 0.78 & 0.12 \\
			MetaMath & 1.04 & 0.68 & 0.12 \\
			
			Qwen3.5-4B & 1.01 & 0.63 & 0.10\\
			Qwen3.5-9B & 0.85 & 0.55 & 0.09  \\ \bottomrule
		\end{tabular}
		\caption{The associated correlation values and RMSE values for each approach.}
		\label{tab:results}
	\end{table}
	
	The fine-tuned Qwen3.5-9B model attains a Pearson correlation of $0.85$ on the
	difficulty parameter $b$, exceeding both regression baselines---MathBERT
	($0.68$) and MetaMath ($0.75$)---by a clear margin. This corresponds to a
	relative improvement of roughly $15\%$ over the stronger baseline, obtained
	with an RMSE of $0.55$ after the linear development-set correction described in
	Section~\ref{sec:method}. The result indicates that the simulated-respondent
	framing recovers difficulty-relevant signal that the direct
	stimulus-to-parameter regression of the baselines does not capture.
	
	When difficulty is discretized into the five-band scheme of
	Table~\ref{tab:diff_labs}, the model reaches a QWK
	of $0.835$ against the calibrated labels, whereas MetaMath has a QWK of $0.692$ and 
	MathBERT has a QWK of $0.625$.

	\section{Discussion}\label{sec:discussion}
	
	Item difficulty modeling remains one of the most difficult tasks in
	computational psychometrics, and our results reinforce both the promise and the
	limits of the simulated-respondent approach. With a modest amount of
	parameter-efficient fine-tuning, a 9-billion-parameter multimodal model
	reconstructs the difficulty parameter at a fidelity (Pearson $0.85$,
	QWK $0.835$) that comfortably exceeds strong text-based regression baselines,
	and it does so for items that mix textual and visual stimuli within a single
	model and a single input representation (Figure~\ref{fig:im-combine}). This
	suggests that the central idea---training the model to replicate the systematic
	pattern of option choices across a discretized range of abilities and then
	reading item parameters off the resulting curves---transfers cleanly to the
	multimodal setting without changes to the underlying architecture.
	
	The most informative aspect of our results is the uneven recovery across
	parameters. Recovering the guessing parameter $c$ at a correlation of $0.48$,
	where the regression baselines are very low by comparison, is arguably the clearest evidence
	that the approach captures genuine response behavior rather than surface
	features of the stimulus: a model that merely reads the question cannot easily
	infer how often low-ability students will be drawn to a particular distractor,
	whereas a model trained to imitate those students can. Conversely, the weak
	recovery of the discrimination parameter $a$ indicates that slope information is
	partially lost, most plausibly through the discretization of ability into
	intervals and through the regression-based correction applied to the raw
	estimates. Finer-grained ability binning, at additional computational cost, is
	a natural avenue for improving slope recovery.
	
	Several design choices bound these results and point to concrete extensions.
	First, to avoid the training instabilities reported in the literature, we
	applied LoRA only to the Gated Attention components of the hybrid Qwen3.5
	architecture, leaving the Gated DeltaNet layers untouched; this adapts roughly
	a quarter of the layers that a conventional transformer would expose to
	fine-tuning. A better understanding of how to adapt the DeltaNet layers stably
	would likely improve recovery, particularly of the parameters that are weakest
	under the present regime. Second, every arrow in our pipeline introduces
	approximation error---the mapping from empirical responses to IRT parameters
	assumes that the 3PL and MCM are good fits, and the discrete reconstruction
	replaces continuous curves with binned expectations---so the reported
	correlations are best read as lower bounds on what the framing can achieve under
	tighter modeling assumptions. 
	
	\section*{Acknowledgments}
	
	\bibliography{references}
	\appendix
	
\end{document}